\documentclass[letterpaper, 10 pt, conference]{ieeeconf}  

\IEEEoverridecommandlockouts                              

\title{\LARGE \bf
DEEP-SEA: Deep-Learning Enhancement for Environmental Perception in Submerged Aquatics}

\author{Shuang Chen\textsuperscript{1}, Ronald Thenius\textsuperscript{2}, Farshad Arvin\textsuperscript{1} and Amir Atapour-Abarghouei\textsuperscript{1}
\thanks{This work was partially supported by Horizon Europe BioDiMoBot [101181363] project.}
\thanks{\textsuperscript{1}S. Chen, F. Arvin and A. Atapour-Abarghouei are with the Department of Computer Science, Durham University, Durham, UK
        {\tt\small shuang.chen@durham.ac.uk}}%
\thanks{\textsuperscript{2}Ronald Thenius is with the Institute of Biology, University of Graz, Graz, Austria.%
}
}
\usepackage{graphicx}
\usepackage{amsmath}
\usepackage{amssymb}
\usepackage{xcolor}
\usepackage{hyperref}
\usepackage[utf8]{inputenc}
\usepackage[export]{adjustbox}
\usepackage{multirow}
\usepackage{makecell}
\usepackage{booktabs}
\newcommand\T{\rule{0pt}{2.9ex}}       
\newcommand\B{\rule[-1.2ex]{0pt}{0pt}} 
\usepackage{url}
\begin{document}
\maketitle
\thispagestyle{empty}
\pagestyle{empty}


\begin{abstract}
Continuous and reliable underwater monitoring is essential for assessing marine biodiversity, detecting ecological changes and supporting autonomous exploration in aquatic environments. Underwater monitoring platforms rely on mainly visual data for marine biodiversity analysis, ecological assessment and autonomous exploration. However, underwater environments present significant challenges due to light scattering, absorption and turbidity, which degrade image clarity and distort colour information, which makes accurate observation difficult. To address these challenges, we propose DEEP-SEA, a novel deep learning-based underwater image restoration model to enhance both low- and high-frequency information while preserving spatial structures. The proposed Dual-Frequency Enhanced Self-Attention Spatial and Frequency Modulator aims to adaptively refine feature representations in frequency domains and simultaneously spatial information for better structural preservation. Our comprehensive experiments on EUVP and LSUI datasets demonstrate the superiority over the state of the art in restoring fine-grained image detail and structural consistency. By effectively mitigating underwater visual degradation, DEEP-SEA has the potential to improve the reliability of underwater monitoring platforms for more accurate ecological observation, species identification and autonomous navigation.
\end{abstract}
\section{Introduction} \label{s:intro}

Robotics technology has become an essential tool in advancing underwater biological research to enable large-scale long-term ecological monitoring. Integrated with autonomous and remotely operated platforms, these underwater monitoring solutions can explore complex underwater environments, detect marine organisms and collect critical visual data to support biodiversity studies, behavioral analysis and environmental monitoring. By reducing human intervention, robotic systems offer increased efficiency, safety and accessibility in deep-sea and remote marine ecosystems.

Underwater data collection and biodiversity monitoring require advanced technologies capable of continuous operation for extended periods without interruption. To address this challenge, approaches such as biohybrid entities for environmental monitoring \cite{thenius2021biohybrid} integrate living organisms with technological components in a symbiotic manner. These biohybrid entities possess distinctive capabilities in energy harvesting, sensing and actuation, which allows them to seamlessly adapt to their operational environment.
Furthermore, they are autonomous, resilient and environmentally friendly due to the incorporation of adapted life forms and biodegradable materials. The novel monitoring technology relies heavily on visual observation, processing vast amounts of image data captured from living organisms such as Zebra mussels~\cite{Niharika2024} and Daphnia~\cite{rajewicz2023organisms} for environmental analysis.

\begin{figure}
    \centering
    \includegraphics[width=0.47\textwidth]{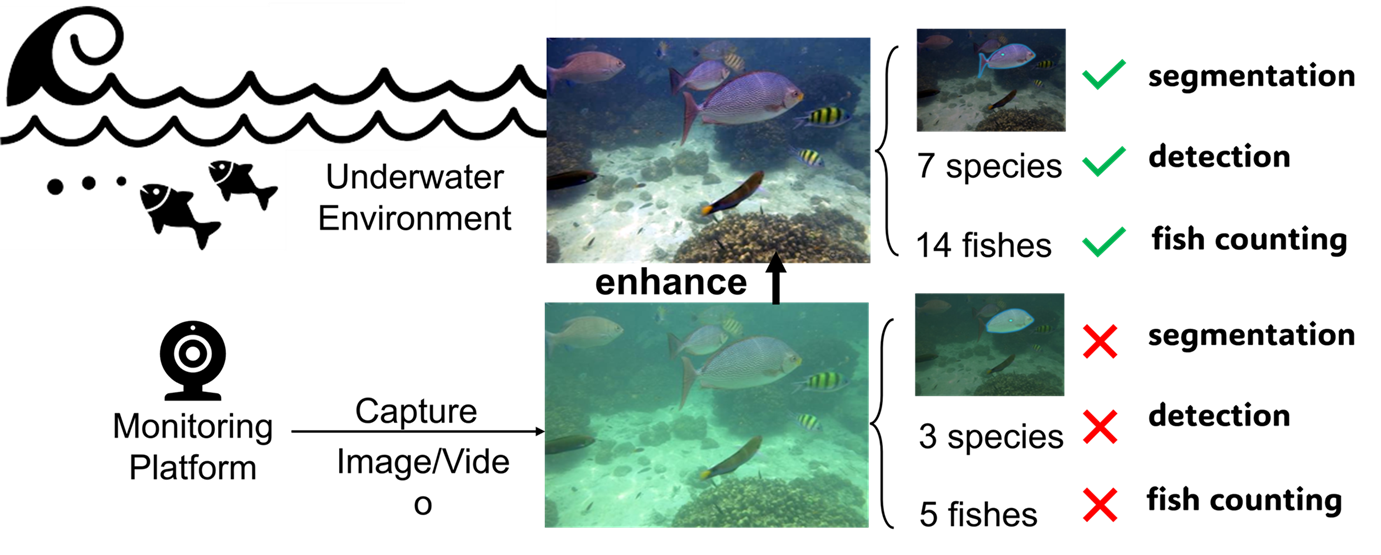}
    \vspace{-0.4cm}
    \caption{An underwater monitoring platform captures visual data that is often degraded due to poor visibility, colour distortion and lighting conditions. Such degraded data negatively impacts downstream tasks and makes marine research, object detection and ecological monitoring challenging. By applying enhancement techniques to restore clarity and colour fidelity, the data becomes significantly more valuable and suitable for downstream tasks, maximising the effectiveness of underwater monitoring efforts.}
    \vspace{-0.4cm}
    \label{fig:teaser}
\end{figure}

However, underwater environments present highly complex and challenging conditions for visual perception. Light scattering, absorption and turbidity severely degrade the quality of visual information captured by the sensors, which can result in reduced visibility, colour distortion and blurring. These degradations are often caused by different types of noise: low-frequency noise, such as haze and colour attenuation, and high-frequency noise, including fine-grained turbulence and texture loss. Such visual degradation significantly impairs object detection, classification and tracking (shown in Fig.~\ref{fig:teaser}), which consequently makes underwater observation and ecological analysis more challenging.

Computer vision methods play a vital role in addressing these challenges, and provide advanced solutions to process and interpret degraded visual data~\cite{chen2025deep}. Techniques such as image enhancement enable more accurate object detection and species identification. Additionally, deep learning-based computer vision models can effectively learn patterns from distorted data to enhance the ability to recognise and track marine organisms even in visually degraded environments. To achieve this, it is important to mitigate degradation present in both low- and high-frequency domains. Although some research has introduced frequency-based features to facilitate restoration~\cite{zhao2024wavelet,guo2024underwater,zhao2024toward}, two key challenges remain: 1) High- and low-frequency components are inherently interwoven, which makes distinguishing their contributions to degradation difficult. High-frequency components capture edges and structural details, while low-frequency components represent background colour and smooth variations. Since both change simultaneously, accurately identifying the dominant degradation factor is challenging. For example, when removing high-frequency noise from a low-frequency background, the restoration process must ensure that the underlying structure remains intact rather than being overly smoothed or distorted. 2) Frequency-domain transformations, such as the Fast Fourier Transform (FFT), primarily focus on global frequency distributions, often neglecting spatial locality~\cite{atapour2016back}. This can result in a loss of structural integrity and spatial coherence in the restored image. The challenge lies in integrating spatial information with frequency-based processing to enhance restoration quality while preserving fine-grained details for accurate visual interpretation.

To address these challenges, we propose DEEP-SEA to restore underwater images by simultaneously recovering low- and high-frequency information, without any loss of spatial awareness, to improve the fine-grained visual detail, which is detailed in Sec:~\ref{sec:method:model}. In the DEEP-SEA, we introduce a Dual-Frequency Enhanced Self-Attention (DFESA) module to adaptively enhance feature representation in low and high frequency features. To further improve the representation learning capacity across multiple scales within the network, we introduce a Spatial and Frequency Modulator (SFM) to enhance both channel-wise and pixel-wise feedforward features.
Our contributions are thus summarised as:
\begin{itemize}
    \item  
    We propose DEEP-SEA, a novel underwater image restoration model to process images captured from our biohybrid platform with a low-power STM32 MCU and Raspberry Pi module for long-term underwater image capture to address underwater image quality issues.
    \item We propose a novel Dual-Frequency Enhanced Self-Attention mechanism  that separately processes low- and high-frequency components to improve feature representation and restore structural details.
    \item We propose a novel Spatial-Frequency Modulation to integrate spatial and frequency cues, ensuring better feature discrimination, noise reduction, and colour correction in underwater images.
\end{itemize}
Our DEEP-SEA outperforms the state-of-the-art methods across both LSUI, and EUVP real-world underwater image datasets. The code is publicly available: {\url{https://github.com/ChrisChen1023/DEEP-SEA}.

\section{Related Work} 
\label{s:related_work}

\begin{figure*}[ht]
\begin{center}
\includegraphics[width=17cm]{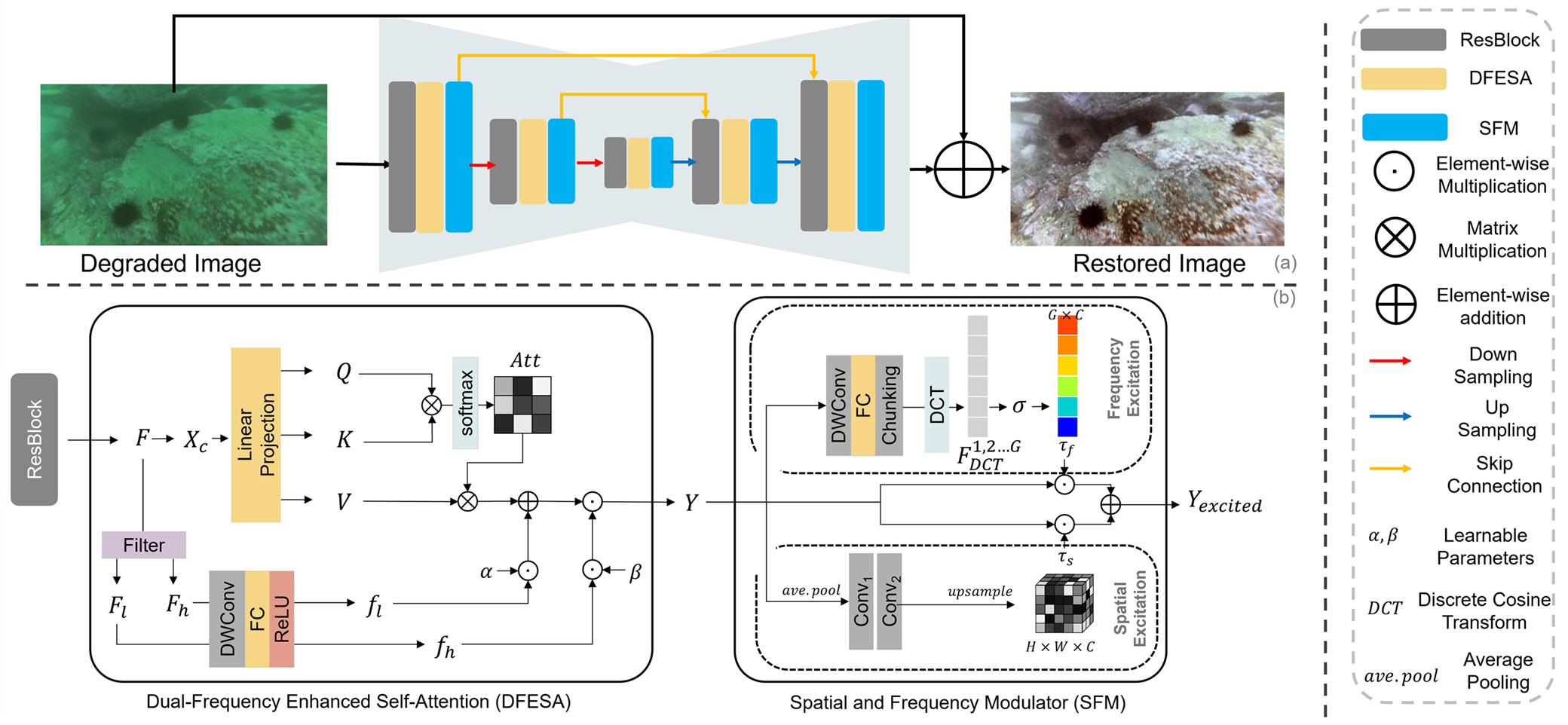}
\end{center}
\vspace{-0.6cm}
   \caption{(a) Architecture Overview; (b) The Dual-Frequency Enhanced Self-Attention module (left) and the Spatial and Frequency Modulator (right).}
   \vspace{-0.5cm}
\label{fig:overview}
\end{figure*}

\subsection{Underwater Environmental Monitoring}

Underwater environmental monitoring has traditionally relied on conventional sensor networks and manual sampling methods to assess water quality, pollutant levels and ecological changes \cite{banna2014online}. While these approaches provide high precision, they often suffer from limitations such as high operational costs, energy demands and infrequent data collection. Recent advancements in biohybrid systems, which integrate living organisms as natural biosensors, offer a promising alternative by leveraging biological responses to environmental stimuli for real-time monitoring \cite{rajewicz2023organisms}. For instance, mussels have been employed as bioindicators due to their valve movements in response to pollutants \cite{dzierzynska2019scares}, and microbial fuel cells have been explored for both power generation and ecosystem monitoring through microbial metabolic activity \cite{walter2020lab}. Such biohybrid approaches are viable energy-efficient solutions for continuous in-situ monitoring, which can contribute to early-warning systems and long-term ecological assessments \cite{thenius2021biohybrid}. However, accurately extracting and interpreting biological responses from underwater imagery remains a challenge due to poor visibility, low contrast and noise. In this work, we focus on developing an effective deep learning-based image enhancement technique to improve the reliability of visual biohybrid monitoring systems.

\subsection{Underwater Image Enhancement}
Underwater image restoration has significantly advanced over time. Early approaches relied on prior knowledge and handcrafted features~\cite{chiang2011underwater,akkaynak2019sea}. However, these traditional methods often fail in diverse underwater environments. 

With the rise of deep learning, more effective restoration techniques have emerged. These methods leverage physical models~\cite{wang2019underwater} and are capable of adapting to various underwater conditions to improve restoration performance. ~\cite{li2020underwater} proposes an underwater image enhancement convolutional model based on underwater scene priors. \cite{cong2023pugan} explores a CNN-based Generative Adversarial Network (GAN) for generating high-quality restored images. However, their performance is constrained by the unstable training of GAN and the limited receptive field of CNNs.

To solve these limitations, researchers have explored alternative architectures. \cite{guo2024underwater} introduce large kernel convolutions to expand receptive fields. \cite{peng2023u} leverages Transformers to capture long-range dependencies for better feature representation learning. \cite{khan2024phaseformer} further improve structural restoration by proposing a phase-based self-attention mechanism.

More recently, researchers have recognised the importance of frequency-domain information in underwater image enhancement. \cite{zhao2024wavelet} propose a wavelet-based method that integrates Fourier information for improved visual quality of the results.   
\cite{guo2024underwater} utilise frequency features to capture complex patterns in shallow layers and demonstrate the potential of frequency-based techniques in underwater image restoration. \cite{zhao2024toward} introduce a spatial-frequency interaction mechanism using cross-connections, leveraging FFT to refine spectral representations in the frequency domain for more effective underwater image restoration.

\section{Methodology} \label{s:method}

Our overall approach involves receiving degraded images from an underwater observation platform (Section \ref{sec:method:robot}) to be processed and enhanced by our underwater image restoration model (Section \ref{sec:method:model}
).

\subsection{Biohybrid Monitoring Platform}
\label{sec:method:robot}

The primary contribution of this paper is the design of a learning-based image restoration model to process images captured under water -- e.g., our previously developed underwater observation system~\cite{thenius2021biohybrid} (shown in Fig.~\ref{fig:RC}). Any platform designed for this task must autonomously function underwater for prolonged periods. At its core, the device employs a low-power STM32 MCU, which can control the peripherals (i.e. Raspberry Pi, sensor modules, Micro SD and wireless communication components, shown in Fig.~\ref{fig:RC-Arch}). Furthermore, the STM32 not only offers multiple onboard communication interfaces (e.g., I2C, SPI, and UART) for data exchange between the peripherals but also supports various operational states (including run, standby, stop and deep sleep modes) to optimise the MCU's power efficiency.

The platform gathers a substantial amount of sensor data for long-term autonomous underwater measurement. The data  is stored in a non-volatile micro SD card. However, the focus of this work is only on underwater images captured and recorded by the Raspberry Pi module. As visually interpreting underwater environments is challenging, these images often suffer from issues such as low contrast, turbidity-induced blurring and colour distortion, which can hinder reliable biological observations. Therefore, we focus on image restoration techniques to improve the clarity and usability of the captured visual data for biohybrid monitoring.

\begin{figure}
    \centering
    \includegraphics[width=0.43\textwidth]{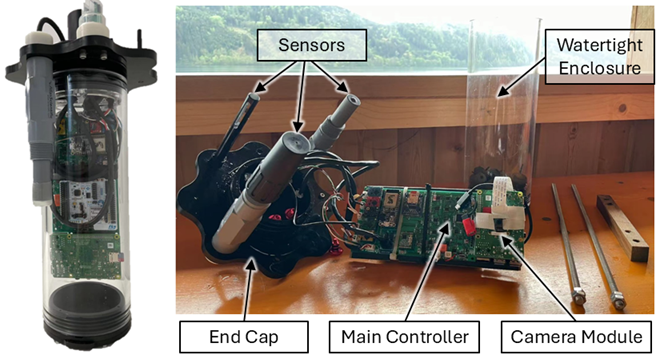}
    \vspace{-0.2cm}
    \caption{The submerged robot and its components which have been developed for long-term underwater monitoring.}
    \vspace{-0.6cm}
    \label{fig:RC}
\end{figure}
\vspace{-0.5cm}
\begin{figure}
    \centering
    \includegraphics[width=0.4 \textwidth]{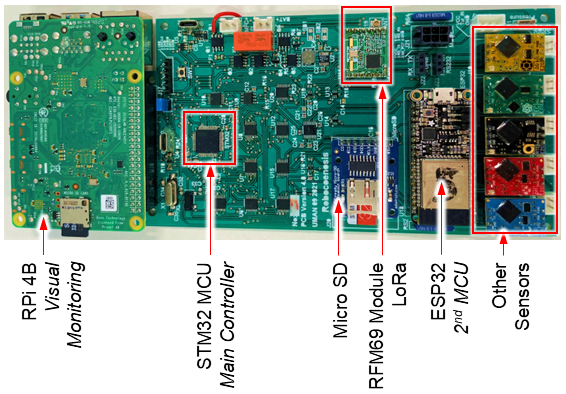}
    \vspace{-0.3cm}
    \caption{The developed core electronic board for long-term monitoring.}
    \vspace{-0.6cm}
    \label{fig:RC-Arch}
\end{figure}
\vspace{0.2cm}

\subsection{Image restoration model architecture}
\label{sec:method:model}

Our primary approach provides a robust end-to-end architecture for underwater image restoration. The proposed architecture adopts a U-Net structure, known for its efficacy in extracting multi-scale features, making it well-suited to reconstructing clean high-quality images from severely degraded inputs. In each block of our U-Net, we use a ResBlock in~\cite{cui2023image}, which takes advantage of average pooling to extract the DC component, which is crucial for reconstructing global consistency. Followed by our novel DFESA module (Section~\ref{sec:method:model:DFESA}), which adaptively enhances feature representations in different frequency components. In the end, we introduce SFM (Section~\ref{sec:method:model:SFM}) to boost feature discrimination by fusing spatial and frequency cues.

\subsubsection{Dual-Frequency Enhanced Self-Attention}
\label{sec:method:model:DFESA}
Although self-attention excels at learning feature representations, it struggles to effectively capture degraded information presented in low- and high-frequency forms, as self-attention is inherently less sensitive to frequency-specific patterns. Theoretically, low-frequency and high-frequency features represent two heterogeneous modalities~\cite{zhou2023xnet}, with the former emphasising sharp transitions, edges, or object boundaries, and the latter capturing consistent, smooth regions such as background colour patterns within an image. Directly integrating them into a single self-attention mechanism without distinction could weaken feature representation capabilities and reduce of the effectiveness of the learning process.

As shown in Fig.~\ref{fig:overview}, our DFESA module separately processes low- and high-frequency information to enhance feature learning. Specifically, within each level of the U-Net-shaped architecture, after extracting feature maps $F$ using ResBlocks, we perform dual-frequency feature fusion via channel-wise self-attention.
First, given the input feature $F$:
\begin{equation}
\begin{gathered}
{X}_{c} = LN(F),\\
Q = W_{q} X_{c}, \quad K = W_{k} X_{c}, \quad V = W_{v} X_{c},
\end{gathered}
\end{equation}
where $LN$ is layer normalisation, ${W_{q,k,v}}$ are learnable weight matrices that project the normalised $X_c$ into queries $Q$, keys $K$, and values $V$ for self-attention. 
The attention scores are obtained through a scaled dot-product operation followed by a softmax function $\varphi(\cdot)$.  The output $\hat{X}_{c}$ is subsequently obtained:
\begin{equation}
\begin{gathered}
\text{Att}_{c}(X_{c}) = \varphi\left(\frac{QK^T}{\sqrt{d}}\right),\\
\hat{X}_{c} = V \odot \text{Att}_{c}(X_{c}),
\end{gathered}
\end{equation}
where $\sqrt{d}$ is the scaling factor to stabilise gradients, $\odot$ denotes element-wise multiplication to integrate attention scores with value features.

To enhance $\hat{X}_{c}$ in dual-frequency perspectives, we decompose input feature $F$ into low- and high-frequency components $F_l$ and $F_h$ with a learnable frequency filter \cite{cui2023image}. $F_l$ and $F_h$  are then processed by a depth-wise convolution followed by a fully-connected layer with a $ReLU$ activation, which will produce low- and high-frequency factors $f_l$ and $f_h$:
\begin{equation}
\begin{gathered}
f_l = {ReLU}(FC(DWConv(F_l))), \\
f_h = {ReLU}(FC(DWConv(F_h))),
\end{gathered}
\end{equation}
where $DWConv$ and $FC$ are the depth-wise convolution and fully-connected layer, respectively.

The self-attention mechanism is refined using the factors $f_l$ and $f_h$. To adaptively refine the weight for different frequency information, we implement $\alpha$ and $\beta$ as two learnable parameters. First, the $f_l$ enhances $\hat{X}_{c}$ on low frequency domain through pixel-wise addition. Then, $f_h$ highlights fine-grained details via pixel-wise multiplication:
\begin{equation}
\begin{gathered}
\hat{X}_{c}^{'} = \hat{X}_{c} + \alpha f_l,\\
Y = \hat{X}_{c}^{'} \odot \beta f_h,
\end{gathered}
\end{equation}
where $\odot$ denotes element-wise multiplication. $Y$ will be fed into the following Spatial and Frequency Modulator module. 
\begin{figure*}[t]
    \centering
    \begin{minipage}[b]{1\linewidth}
        \begin{minipage}[b]{.12\linewidth}
            \centering
            \includegraphics[width=\linewidth]{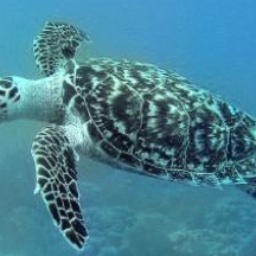}
        \end{minipage}
        \begin{minipage}[b]{.12\linewidth}
            \centering
            \includegraphics[width=\linewidth]{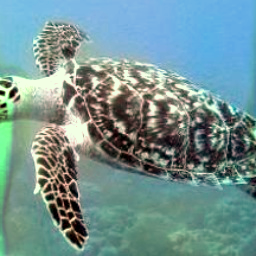}
        \end{minipage}
        \begin{minipage}[b]{.12\linewidth}
            \centering
            \includegraphics[width=\linewidth]{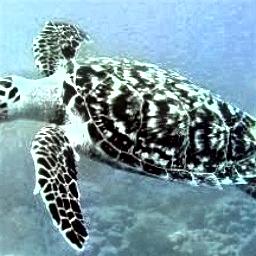}
        \end{minipage}
        \begin{minipage}[b]{.12\linewidth}
            \centering
            \includegraphics[width=\linewidth]{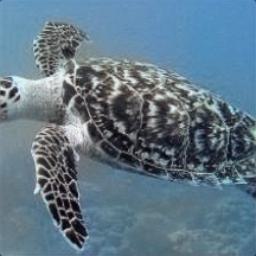}
        \end{minipage}
        \begin{minipage}[b]{.12\linewidth}
            \centering
            \includegraphics[width=\linewidth]{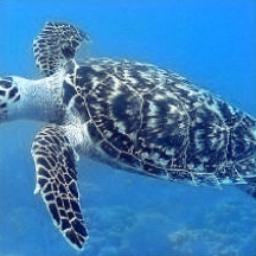}
        \end{minipage}
        \begin{minipage}[b]{.12\linewidth}
            \centering
            \includegraphics[width=\linewidth]{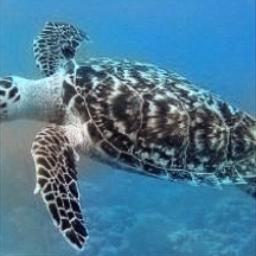}
        \end{minipage}
        \begin{minipage}[b]{.12\linewidth}
            \centering
            \includegraphics[width=\linewidth]{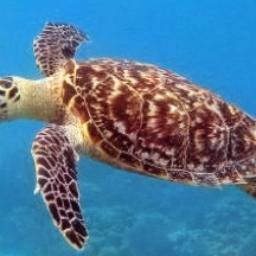}
        \end{minipage}
        \begin{minipage}[b]{.12\linewidth}
            \centering
            \includegraphics[width=\linewidth]{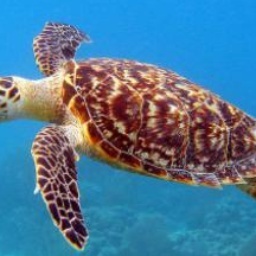}
        \end{minipage}
    \end{minipage}

    \begin{minipage}[b]{1\linewidth}
        \begin{minipage}[b]{.12\linewidth}
            \centering
            \includegraphics[width=\linewidth]{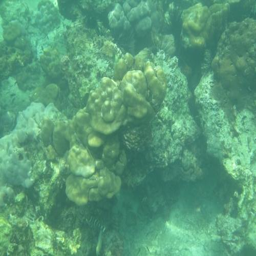}
        \end{minipage}
        \begin{minipage}[b]{.12\linewidth}
            \centering
            \includegraphics[width=\linewidth]{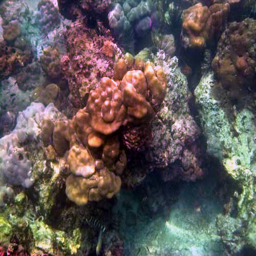}
        \end{minipage}
        \begin{minipage}[b]{.12\linewidth}
            \centering
            \includegraphics[width=\linewidth]{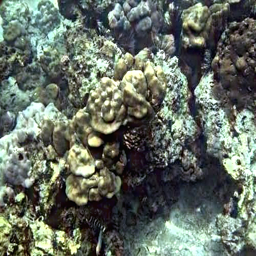}
        \end{minipage}
        \begin{minipage}[b]{.12\linewidth}
            \centering
            \includegraphics[width=\linewidth]{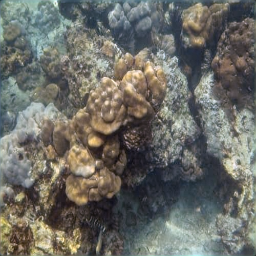}
        \end{minipage}
        \begin{minipage}[b]{.12\linewidth}
            \centering
            \includegraphics[width=\linewidth]{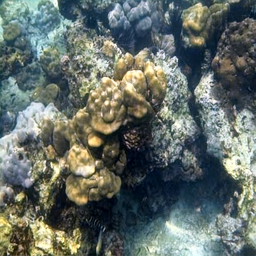}
        \end{minipage}
        \begin{minipage}[b]{.12\linewidth}
            \centering
            \includegraphics[width=\linewidth]{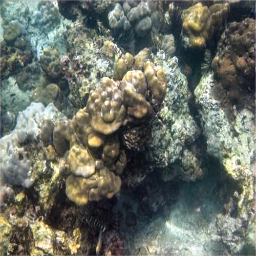}
        \end{minipage}
        \begin{minipage}[b]{.12\linewidth}
            \centering
            \includegraphics[width=\linewidth]{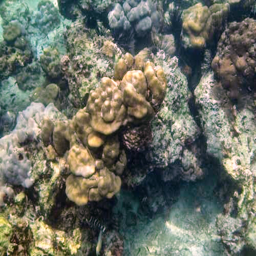}
        \end{minipage}
        \begin{minipage}[b]{.12\linewidth}
            \centering
            \includegraphics[width=\linewidth]{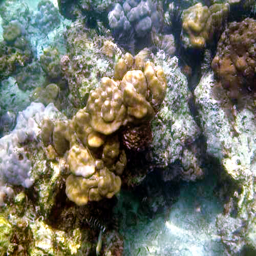}
        \end{minipage}
    \end{minipage}

    \begin{minipage}[b]{1\linewidth}
        \begin{minipage}[b]{.12\linewidth}
            \centering
            \includegraphics[width=\linewidth]{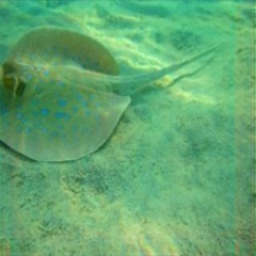}
        \end{minipage}
        \begin{minipage}[b]{.12\linewidth}
            \centering
            \includegraphics[width=\linewidth]{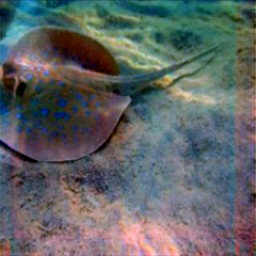}
        \end{minipage}
        \begin{minipage}[b]{.12\linewidth}
            \centering
            \includegraphics[width=\linewidth]{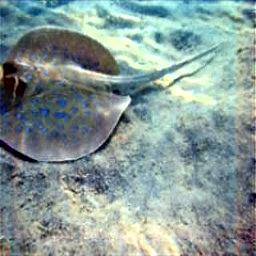}
        \end{minipage}
        \begin{minipage}[b]{.12\linewidth}
            \centering
            \includegraphics[width=\linewidth]{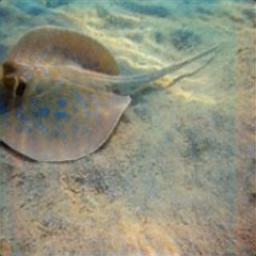}
        \end{minipage}
        \begin{minipage}[b]{.12\linewidth}
            \centering
            \includegraphics[width=\linewidth]{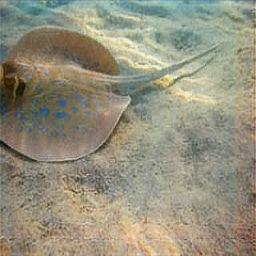}
        \end{minipage}
        \begin{minipage}[b]{.12\linewidth}
            \centering
            \includegraphics[width=\linewidth]{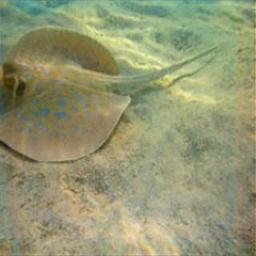}
        \end{minipage}
        \begin{minipage}[b]{.12\linewidth}
            \centering
            \includegraphics[width=\linewidth]{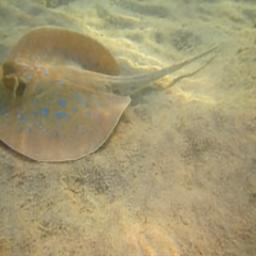}
        \end{minipage}
        \begin{minipage}[b]{.12\linewidth}
            \centering
            \includegraphics[width=\linewidth]{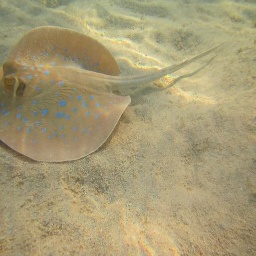}
        \end{minipage}
    \end{minipage}

    \begin{minipage}[b]{1\linewidth}
        \begin{minipage}[b]{.12\linewidth}
            \centering
            \includegraphics[width=\linewidth]{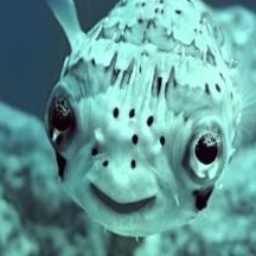}
        \end{minipage}
        \begin{minipage}[b]{.12\linewidth}
            \centering
            \includegraphics[width=\linewidth]{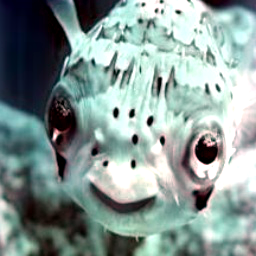}
        \end{minipage}
        \begin{minipage}[b]{.12\linewidth}
            \centering
            \includegraphics[width=\linewidth]{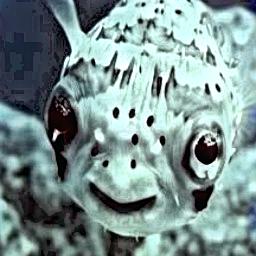}
        \end{minipage}
        \begin{minipage}[b]{.12\linewidth}
            \centering
            \includegraphics[width=\linewidth]{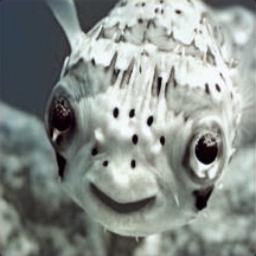}
        \end{minipage}
        \begin{minipage}[b]{.12\linewidth}
            \centering
            \includegraphics[width=\linewidth]{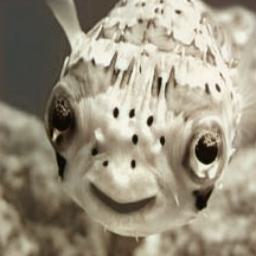}
        \end{minipage}
        \begin{minipage}[b]{.12\linewidth}
            \centering
            \includegraphics[width=\linewidth]{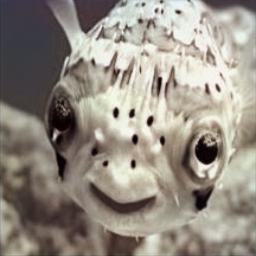}
        \end{minipage}
        \begin{minipage}[b]{.12\linewidth}
            \centering
            \includegraphics[width=\linewidth]{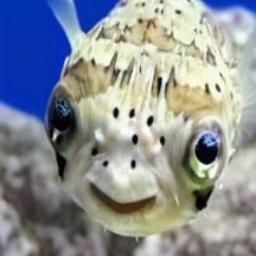}
        \end{minipage}
        \begin{minipage}[b]{.12\linewidth}
            \centering
            \includegraphics[width=\linewidth]{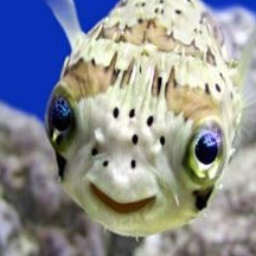}
        \end{minipage}
    \end{minipage}

    \begin{minipage}[b]{1\linewidth}
        \begin{minipage}[b]{.12\linewidth}
            \centering
            \centerline{\includegraphics[width=\linewidth]{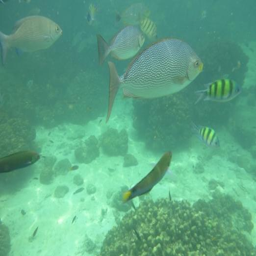}}
            \centerline{(a) Input}\medskip
        \end{minipage}
        \begin{minipage}[b]{.12\linewidth}
            \centering
            \centerline{\includegraphics[width=\linewidth]{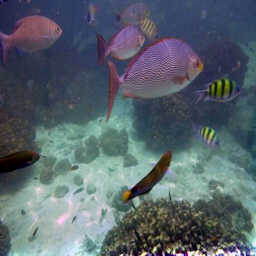}}
            \centerline{(b) ADPCC}\medskip
        \end{minipage}
        \begin{minipage}[b]{0.12\linewidth}
            \centering
            \centerline{\includegraphics[width=\linewidth]{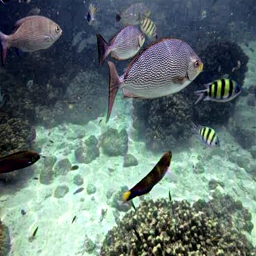}}
            \centerline{(c) WWPF}\medskip
        \end{minipage}
        \begin{minipage}[b]{.12\linewidth}
            \centering
            \centerline{\includegraphics[width=\linewidth]{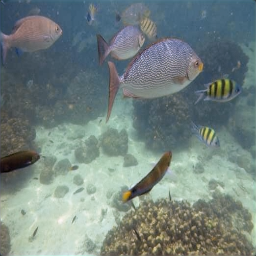}}
            \centerline{(d) GUPDM}\medskip
        \end{minipage}
        \begin{minipage}[b]{.12\linewidth}
            \centering
            \centerline{\includegraphics[width=\linewidth]{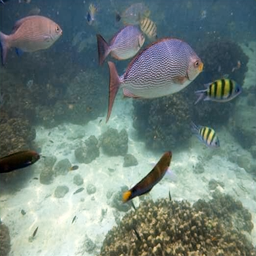}}
            \centerline{(e) Semi-UIR}\medskip
        \end{minipage}
        \begin{minipage}[b]{.12\linewidth}
            \centering
            \centerline{\includegraphics[width=\linewidth]{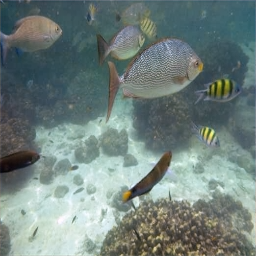}}
            \centerline{(f) URSCT}\medskip
        \end{minipage}
        \begin{minipage}[b]{.12\linewidth}
            \centering
            \centerline{\includegraphics[width=\linewidth]{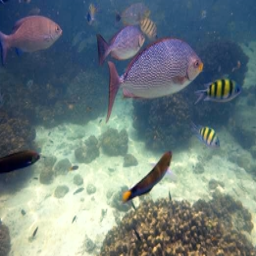}}
            \centerline{(g) Ours}\medskip
        \end{minipage}
        \begin{minipage}[b]{0.12\linewidth}
            \centering
            \centerline{\includegraphics[width=\linewidth]{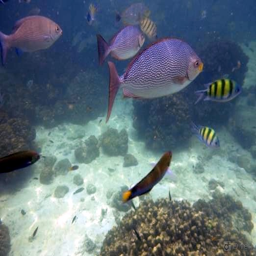}}
            \centerline{(h) Reference}\medskip
        \end{minipage}
    \end{minipage}

    \vspace{-0.5cm}
    \caption{Samples for visual comparison showcase how the proposed method achieves more plausible color restoration and clearer structural details by effectively capturing both low- and high-frequency information, leading to superior image quality.} 
    \vspace{-0.4cm}
    \label{fig:comp}
\end{figure*}
\subsubsection{Spatial and Frequency Modulator}
\label{sec:method:model:SFM}
In DFESA, features are implicitly enriched through a combination of low- and high-frequency indicators. To further explicitly enhance feature representations in the spatial and frequency domains, we introduce the SFM. This module separately models both frequency and spatial information to refine the feature for a more comprehensive representation.

\noindent\textbf{Frequency Excitation Branch:} As shown in Fig.~\ref{fig:overview}, the frequency excitation branch is designed to capture spectral dependencies across different channels to allow the model to focus on frequency-specific features. Specifically, we first apply a depthwise convolution, followed by a FC layer, which refines receptive fields while maintaining computational efficiency.
Next, we adopt the multi-spectral channel attention mechanism~\cite{qin2021fcanet} to extract a frequency-aware channel attention vector. Formally, the input $Y$ is split into multiple channel-wise chunks to allow for separate frequency processing. Each chunk undergoes a 2D Discrete Cosine Transform $D(\cdot)$) to obtain a compressed spectral representation $F_{\text{DCT}}$:
\begin{equation}
F_{\text{DCT}}^g = \mathcal{D}(X_g), \quad g = 1,2,...,G,
\end{equation}
where $X_g$ represents the $g$-th channel group and $G$ is the total number of groups.
To generate the frequency attention vector $\tau_f$, we aggregate $F_{\text{DCT}}^g$ and pass it through a projection  $\sigma(\cdot)$, implemented by fully-connected layers, ReLU and Sigmoid activations:
\begin{equation} 
\tau_f = \sigma(F_{\text{DCT}}).
\end{equation}

\noindent\textbf{Spatial Excitation Branch:} we follow the spatial attention mechanism from \cite{chen2024hint}. Specifically, instead of directly applying convolutions to extract spatial relations, we first partially downscale the input feature map to preserve global spatial dependencies while reducing computational overhead. Formally, given the input feature $Y$, we apply an average pooling to half downscale $Y$ to $Y_s$, followed by two consecutive {$Convolution-LayerNorm-ReLU$} blocks (named $Conv$):
\begin{equation}
I_s = \text{Conv}_2(\text{Conv}_1(Y_s)),
\end{equation}
where \(\text{Conv}_1 \) and \(\text{Conv}_2 \) are two convolutional layers responsible for extracting spatial dependencies. After upsampling $Y_s$, we obtain the final spatial indicator $\tau_s$.
%
\noindent\textbf{Final Feature Fusion:} The final feature excitation is achieved by integrating both spatial and frequency indicators for a more comprehensive feature refinement. The frequency excitation vector $\tau_f$ is applied to the feature $Y$ via channel-wise multiplication to further enhance spectral dependencies by adaptively amplifying or suppressing specific frequency components. 
Similarly, the spatial attention tensor $\tau_s$ is applied to $Y$ but through pixel-wise multiplication to refine spatial representations by dynamically adjusting feature importance across different locations. After fusing these refined representations, the model effectively captures both frequency- and spatial-domain information:
\begin{equation}
Y_{\text{excited}} = Sum(\tau_f \cdot Y, \tau_s \cdot Y),
\end{equation}

\subsubsection{Loss Function}
To preserve pixel-level details, we use \textit{L1} loss, which is less sensitive to outliers than \textit{L2} loss and effectively retains structural information in reconstructed images~\cite{zhao2016loss}. Beyond spatial-domain supervision, we apply FFT to compute loss in the frequency domain. This dual-domain approach ensures the model captures both fine-grained details for sharp textures and low-frequency components to maintain global illumination consistency. We followed \cite{cho2021rethinking,tu2022maxim} to integrate side encoders and decoders for training stability and multi-scale feature learning. Given the output image $I_{pred}$ and the ground truth $I_{gt}$, the total loss combines spatial \textit{L1} loss and \textit{FFT} loss with balanced optimisation for high-quality reconstruction and stable convergence.

\begin{equation}
\begin{gathered}
\mathcal{L}_{\text{L1}} = \frac{1}{N} \sum_{i=1}^{N} | I_{\text{pred}}(i) - I_{\text{gt}}(i) |,\\
\mathcal{L}_{\text{freq}} = \frac{1}{N} \sum_{i=1}^{N} \left| \mathcal{F}(I_{\text{pred}})(i) - \mathcal{F}(I_{\text{gt}})(i) \right|,\\
\mathcal{L}_{\text{total}} = \lambda_1 \mathcal{L}_{\text{L1}} + \lambda_2 \mathcal{L}_{\text{freq}},
\end{gathered}
\end{equation}
where \( \mathcal{F} \) denotes the FFT operation. \textit{N} is the number of scales. In our work $N=3$.

\section{Results and Discussion} 
\label{s:exp}

We followed the benchmark~\cite{guo2024underwater} on the widely used LSUI~\cite{peng2023u} and EUVP~\cite{islam2020fast} datasets to evaluate the efficacy of our proposed method on real-world samples. As we aim for images that are free of noise, exhibit clear structures, and maintain semantically plausible patterns, our evaluation focuses on fine-grained details by using PSNR, SSIM and LPIPS, to enable a more comprehensive comparison against state-of-the-art methods. As underwater image restoration focuses on refining existing content rather than generating new elements, we do not use metrics such as FID to evaluate the distribution between restored and real images. 

All experiments are implemented in PyTorch on a single NVIDIA Titan RTX GPU, with a batch size of 4. During training, we randomly crop $256\times 256$ patches with random flipping as data augmentation. The learning rate follows a cosine annealing schedule~\cite{loshchilov2016sgdr}, decaying from  $1e^{-4}$ to $1e^{-6}$. We use the Adam optimiser with $\beta_1=0.9$, $\beta_2=0.999$.
\vspace{-0.2cm}

\begin{table}[h]
\centering
\caption{Quantitative comparison on EUVP, and LSUI.}
\label{tab:comp}
\vspace{-0.2cm}
\adjustbox{width=0.98\linewidth}{
\begin{tabular}{l|ccc|ccc}
\toprule
  & \multicolumn{3}{c|}{EUVP} & \multicolumn{3}{c}{LSUI}  \\ 
 \cmidrule(l){2-7} 
                        \multirow{-2}{*}{Method}  & PSNR     & SSIM    & LPIPS     & PSNR     & SSIM    & LPIPS   \\ \midrule
ROP~\cite{liu2022rank}        & 15.34   & 0.714   & 0.343     & 17.38   & 0.806   & 0.281    \\
ADPCC~\cite{zhou2023underwater}   & 15.20   & 0.692   & 0.349   & 16.20   & 0.763   & 0.299    \\
WWPF~\cite{zhang2023underwater}    & 15.95   & 0.648   & 0.337      & 17.90   & 0.739   & 0.283    \\ \midrule
CLUIE-Net~\cite{li2022beyond}   & 24.85   & 0.844   & 0.186     & 23.57   & 0.864   & 0.175   \\
FUIEGAN~\cite{islam2020fast} & 25.46   & 0.770   & 0.242  & 24.23   & 0.858    & 0.182 \\
TACL~\cite{liu2022twin}        & 20.99   & 0.782   & 0.213  & 22.97   & 0.828   & 0.176     \\
UIE-WD~\cite{ma2022wavelet}     & 17.80   & 0.760   & 0.292    & 19.23   & 0.803   & 0.284     \\
URSCT~\cite{ren2022reinforced}      & 25.74   & 0.855   & 0.180     & 25.87   & 0.883   & 0.146   \\
USUIR~\cite{fu2022unsupervised}      & 21.94   & 0.810   & 0.239     & 23.75   & 0.860   & 0.184     \\
GUPDM~\cite{mu2023generalized}    & 24.79   & 0.847   & 0.184   & 25.33   & 0.877   & 0.150     \\
PUGAN~\cite{cong2023pugan}  & 22.58   & 0.820   & 0.212    & 23.14   & 0.836   & 0.216    \\
Semi-UIR~\cite{huang2023contrastive}    & 24.59   & 0.821   & 0.172     & 25.40   & 0.843   & 0.160    \\
TUDA~\cite{wang2023domain}    & 23.73   & 0.843   & 0.207     & 25.52   & 0.878   & 0.154   \\
U-Transformer~\cite{peng2023u} & 24.99   & 0.829   & 0.238    & 25.15   & 0.838   & 0.221    \\
SFGNet~\cite{zhao2024toward}  & 22.68   & 0.585   & 0.221    & 22.71   & 0.653   & 0.204     \\
URI-PolyKernel~\cite{guo2024underwater}   & \underline{26.42}   & \underline{0.866}   & \underline{0.154}    & \underline{26.55}   & \underline{0.888}   & \underline{0.125}   \\ 
\midrule
Ours             & \textbf{29.27}  & \textbf{0.905}  &  \textbf{0.131}  &  \textbf{28.90}  &  \textbf{0.897} &  \textbf{0.090}    \\ \bottomrule
\end{tabular}
}
\vspace{-0.4cm}
\end{table}

\subsection{Quantitative Results}

Quantitative results on the EUVP and LSUI datasets demonstrate the effectiveness of DEEP-SEA, achieving the highest PSNR and SSIM scores for superior perceptual quality and structural consistency, along with the lowest LPIPS for enhanced visual fidelity. On EUVP, DEEP-SEA surpasses URI-PolyKernel~\cite{guo2024underwater} by 10.8\% in PSNR and 4.5\% in SSIM. Similarly, on LSUI, it achieves a PSNR of 28.90 and SSIM of 0.897, improving PSNR by 8.8\% and SSIM by 1.0\%. These results confirm its robustness across diverse underwater scenarios.

\vspace{-0.3cm}
\begin{figure}[h]
    \centering
    \begin{minipage}{0.83\columnwidth}
        \centering
        \includegraphics[width=\linewidth]{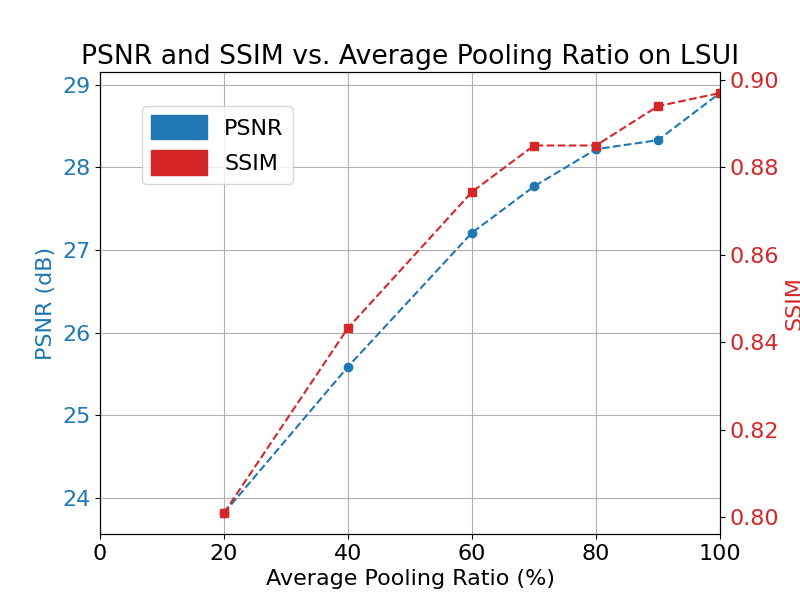}  
    \end{minipage}
    \hfill
    \begin{minipage}{0.83\columnwidth}
        \centering
        \includegraphics[width=\linewidth]{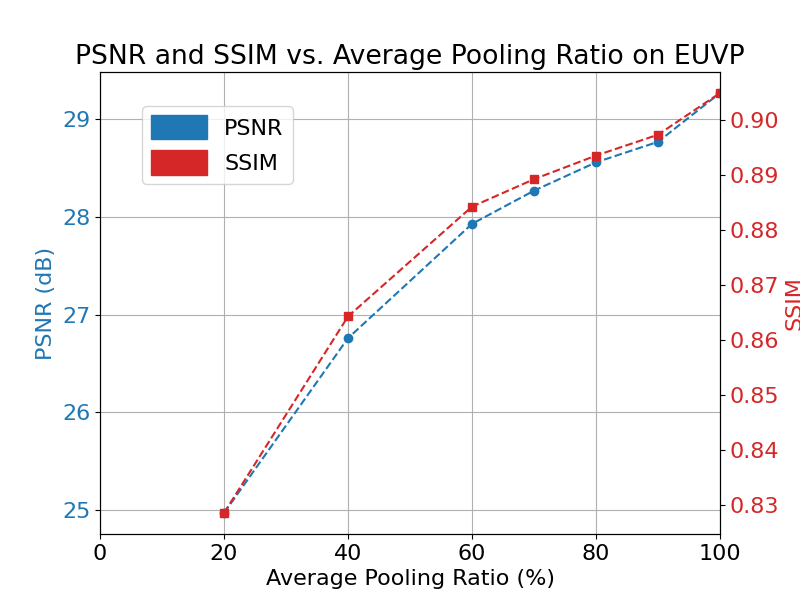}  
    \end{minipage}
    \caption{Impact of average pooling ratio on underwater image restoration performance. 100\% ratio means global average pooling.}
    \vspace{-0.5cm}
    \label{fig:pooling}
\end{figure}
\subsection{Qualitative Results}
Beyond quantitative improvements, our method demonstrates clear visual advantages over the state of the art. As shown in Fig. \ref{fig:comp}, our method effectively restores high-quality images with sharper structures and enhanced visual patterns by mitigating the effects of light scattering, absorption and turbidity. Additionally, our approach fully addresses the problem of colour degradation caused by underwater distortions and recovers the original colours of objects while preserving their structural integrity. Compared to other methods, our results exhibit more vivid textures, more natural colour distribution and clearer object boundaries, all of which highlight the robustness of our enhancement strategy.
\begin{table}[h]
\centering
\caption{Ablation studies of each component.}
\vspace{-0.3cm}
        \resizebox{0.48\textwidth}{!}
        {
		\begin{tabular}{@{\extracolsep{3pt}} c | c c c@{}}
				\hline
					{Variants}  & PSNR$\uparrow$ & SSIM$\uparrow$ & LPIPS$\downarrow$\T\B\\
				\hline
        \textit{\textcolor{black}{Baseline}}  & 26.34  & 0.866  & 0.103   \\
        \textit{\textcolor{black}{Baseline + DFESA}}& 27.68 & 0.880   & 0.094  \\
        \textit{\textcolor{black}{Baseline + DFESA + SFM (ours)}}  &  \textbf{28.90}  &  \textbf{0.897} &  \textbf{0.090}  \B\\
        \hline
        \end{tabular}
        }
\label{tab:ablation}
\vspace{-0.5cm}
\end{table}

\subsection{Ablation Study}
To evaluate the effectiveness of each component in our method, we conduct an ablation study on the LSUI dataset, shown in Tab. \ref{tab:ablation}. The baseline model is a U-Net-shaped architecture with ResBlock, followed by self-attention and MLP feed-forward network. Optimising self-attention to DFESA  improves performance by 5.1\% in PSNR, 1.6\% in SSIM and reduces LPIPS by 8.7\%, demonstrating the benefit of incorporating low- and high- frequency information in self-attention. With SFM, our method gains 5.1\% improvement in PSNR, 1.6\% in SSIM and 4.3\% reduction in LPIPS. These results show the effectiveness of introducing spatial information while fusing it with frequency-excited features.

\subsection{Pooling Ratio Tuning}
We find an interesting and valuable observation that increasing the average pooling ratio in ResBlock leads to a steady improvement in restoration performance, particularly in PSNR. This is likely due to the fact that PSNR is particularly sensitive to global brightness and contrast distortions, which are predominantly influenced by low-frequency degradations; increasing the pooling ratio helps recover these low-frequency components more effectively, which consequently leads to an improvement in the overall restoration fidelity. As shown in Fig~\ref{fig:pooling}, on LSUI, PSNR improves by 5.08 dB and SSIM rises from 0.8010 to 0.897. Similarly, on EUVP, PSNR increases by 4.30 dB and SSIM reaches 0.905.

This trend aligns with the frequency-domain characteristics of underwater images, where degradation primarily affects low-frequency components, such as colour attenuation and global contrast loss. Higher pooling ratios emphasises more global low-frequency noise, enhance global structure consistency and refine colour distributions, which leads to sharper and more stable restorations. This finding highlights the importance of adaptive pooling strategies in achieving a better balance between fine-detail preservation and global feature enhancement for underwater image restoration.

\section{Conclusion} \label{s:conclusion}

In this paper, we have proposed DEEP-SEA, a novel underwater image restoration model that effectively addresses visual degradation in aquatic environments by enhancing both low- and high-frequency information while preserving spatial structures. Our DFESA mechanism adaptively refines frequency-aware feature representations to restore fine-grained textures and global colours consistency. On the other hand, our SFM module integrates spatial cues to enhance structural coherence. Experimental evaluations on EUVP and LSUI datasets demonstrate that DEEP-SEA outperforms state-of-the-art methods with enhanced underwater image clarity and fidelity.

Developing DEEP-SEA and evaluating its efficacy on public datasets is the first step of our project. In the next phase, DEEP-SEA will be embedded into the biohybrid monitoring platform. This integration will benefit other vision-based tasks, such as object recognition, fish counting and marine species monitoring for more reliable biological observations in complex underwater environments.
Additionally, integrating self-supervised learning for DEEP-SEA could reduce reliance on paired training data for generalisation towards more diverse underwater conditions. Further research could also explore domain adaptation techniques to apply DEEP-SEA across different water types, where factors like turbidity, lighting and biological interference vary. Moreover, beyond colour-related degradations, motion blur from sensor movement or ocean currents introduces dynamic distortions. Addressing these varied degradation patterns simultaneously will be a natural next step.

\section{Acknowledgement}
This work was supported by the HEU Project BioDiMoBot, grant agreement No:101181363.
\vspace{-0.2cm}



\bibliographystyle{IEEEtran.bst} 
\bibliography{main}

\begin{thebibliography}{10}
\providecommand{\url}[1]{#1}
\csname url@rmstyle\endcsname
\providecommand{\newblock}{\relax}
\providecommand{\bibinfo}[2]{#2}
\providecommand\BIBentrySTDinterwordspacing{\spaceskip=0pt\relax}
\providecommand\BIBentryALTinterwordstretchfactor{4}
\providecommand\BIBentryALTinterwordspacing{\spaceskip=\fontdimen2\font plus
\BIBentryALTinterwordstretchfactor\fontdimen3\font minus \fontdimen4\font\relax}
\providecommand\BIBforeignlanguage[2]{{%
\expandafter\ifx\csname l@#1\endcsname\relax
\typeout{** WARNING: IEEEtran.bst: No hyphenation pattern has been}%
\typeout{** loaded for the language `#1'. Using the pattern for}%
\typeout{** the default language instead.}%
\else
\language=\csname l@#1\endcsname
\fi
#2}}

\bibitem{thenius2021biohybrid}
R.~Thenius, W.~Rajewicz, J.~C. Varughese, S.~Schoenwetter-Fuchs, F.~Arvin, A.~J. Casson, C.~Wu, B.~Lennox, A.~Campo, G.~J. van Vuuren, \emph{et~al.}, ``Biohybrid entities for environmental monitoring,'' in \emph{The 2021 Conference on Artificial Life}, 2021.

\bibitem{Niharika2024}
N.~Gogoi, N.~Helmer, C.~Wu, R.~Thenius, A.~Casson, T.~Schmickl, and F.~Arvin, ``Biohybrid sensors for underwater monitoring,'' in \emph{IEEE Applied Sensing Conference}, 2024.

\bibitem{rajewicz2023organisms}
W.~Rajewicz, C.~Wu, D.~Romano, A.~Campo, F.~Arvin, A.~J. Casson, G.~J. van Vuuren, C.~Stefanini, \emph{et~al.}, ``Organisms as sensors in biohybrid entities as a novel tool for in-field aquatic monitoring,'' \emph{Bioinspiration \& Biomimetics}, vol.~19, no.~1, p. 015001, 2023.

\bibitem{chen2025deep}
S.~Chen, Y.~He, B.~Lennox, F.~Arvin, and A.~Atapour-Abarghouei, ``Deep learning-enhanced visual monitoring in hazardous underwater environments with a swarm of micro-robots,'' in \emph{IEEE International Conference on Robotics and Automation}, 2025.

\bibitem{zhao2024wavelet}
C.~Zhao, W.~Cai, C.~Dong, and C.~Hu, ``Wavelet-based fourier information interaction with frequency diffusion adjustment for underwater image restoration,'' in \emph{Proceedings of the IEEE/CVF Conference on Computer Vision and Pattern Recognition}, 2024, pp. 8281--8291.

\bibitem{guo2024underwater}
X.~Guo, Y.~Dong, X.~Chen, W.~Chen, Z.~Li, F.~Zheng, and C.-M. Pun, ``Underwater image restoration via polymorphic large kernel cnns,'' \emph{arXiv preprint arXiv:2412.18459}, 2024.

\bibitem{zhao2024toward}
C.~Zhao, W.~Cai, C.~Dong, and Z.~Zeng, ``Toward sufficient spatial-frequency interaction for gradient-aware underwater image enhancement,'' in \emph{IEEE International Conference on Acoustics, Speech and Signal Processing (ICASSP)}, 2024, pp. 3220--3224.

\bibitem{atapour2016back}
A.~Atapour-Abarghouei, G.~P. de~La~Garanderie, and T.~P. Breckon, ``Back to {Butterworth} - a {Fourier} basis for {3D} surface relief hole filling within {RGB-D} imagery,'' in \emph{2016 23rd International Conference on Pattern Recognition (ICPR)}.\hskip 1em plus 0.5em minus 0.4em\relax IEEE, 2016, pp. 2813--2818.

\bibitem{banna2014online}
M.~H. Banna, S.~Imran, A.~Francisque, H.~Najjaran, R.~Sadiq, M.~Rodriguez, and M.~Hoorfar, ``Online drinking water quality monitoring: review on available and emerging technologies,'' \emph{Critical Reviews in Environmental Science and Technology}, vol.~44, no.~12, pp. 1370--1421, 2014.

\bibitem{dzierzynska2019scares}
A.~Dzier{\.z}y{\'n}ska-Bia{\l}o{\'n}czyk, {\L}.~Jermacz, J.~Zielska, and J.~Kobak, ``What scares a mussel? changes in valve movement pattern as an immediate response of a byssate bivalve to biotic factors,'' \emph{Hydrobiologia}, vol. 841, pp. 65--77, 2019.

\bibitem{walter2020lab}
X.~A. Walter, J.~You, J.~Winfield, U.~Bajarunas, J.~Greenman, and I.~A. Ieropoulos, ``From the lab to the field: Self-stratifying microbial fuel cells stacks directly powering lights,'' \emph{Applied energy}, vol. 277, p. 115514, 2020.

\bibitem{chiang2011underwater}
J.~Y. Chiang and Y.-C. Chen, ``Underwater image enhancement by wavelength compensation and dehazing,'' \emph{IEEE Transactions on Image Processing}, vol.~21, no.~4, pp. 1756--1769, 2011.

\bibitem{akkaynak2019sea}
D.~Akkaynak and T.~Treibitz, ``Sea-thru: A method for removing water from underwater images,'' in \emph{The IEEE/CVF Conference on Computer Vision and Pattern Recognition}, 2019, pp. 1682--1691.

\bibitem{wang2019underwater}
K.~Wang, Y.~Hu, J.~Chen, X.~Wu, X.~Zhao, and Y.~Li, ``Underwater image restoration based on a parallel convolutional neural network,'' \emph{Remote sensing}, vol.~11, no.~13, p. 1591, 2019.

\bibitem{li2020underwater}
C.~Li, S.~Anwar, and F.~Porikli, ``Underwater scene prior inspired deep underwater image and video enhancement,'' \emph{Pattern Recognition}, vol.~98, p. 107038, 2020.

\bibitem{cong2023pugan}
R.~Cong, W.~Yang, W.~Zhang, C.~Li, C.-L. Guo, Q.~Huang, and S.~Kwong, ``Pugan: Physical model-guided underwater image enhancement using gan with dual-discriminators,'' \emph{IEEE Transactions on Image Processing}, 2023.

\bibitem{peng2023u}
L.~Peng, C.~Zhu, and L.~Bian, ``U-shape transformer for underwater image enhancement,'' \emph{IEEE Transactions on Image Processing}, 2023.

\bibitem{khan2024phaseformer}
M.~Khan, A.~Negi, A.~Kulkarni, S.~S. Phutke, S.~K. Vipparthi, and S.~Murala, ``Phaseformer: Phase-based attention mechanism for underwater image restoration and beyond,'' \emph{arXiv preprint arXiv:2412.01456}, 2024.

\bibitem{cui2023image}
Y.~Cui, W.~Ren, X.~Cao, and A.~Knoll, ``Image restoration via frequency selection,'' \emph{IEEE Transactions on Pattern Analysis and Machine Intelligence}, vol.~46, no.~2, pp. 1093--1108, 2023.

\bibitem{zhou2023xnet}
Y.~Zhou, J.~Huang, C.~Wang, L.~Song, and G.~Yang, ``Xnet: Wavelet-based low and high frequency fusion networks for fully-and semi-supervised semantic segmentation of biomedical images,'' in \emph{Proceedings of the IEEE/CVF International Conference on Computer Vision}, 2023, pp. 21\,085--21\,096.

\bibitem{qin2021fcanet}
Z.~Qin, P.~Zhang, F.~Wu, and X.~Li, ``Fcanet: Frequency channel attention networks,'' in \emph{Proceedings of the IEEE/CVF international conference on computer vision}, 2021, pp. 783--792.

\bibitem{chen2024hint}
S.~Chen, A.~Atapour-Abarghouei, and H.~P. Shum, ``Hint: High-quality inpainting transformer with mask-aware encoding and enhanced attention,'' \emph{IEEE Transactions on Multimedia}, 2024.

\bibitem{zhao2016loss}
H.~Zhao, O.~Gallo, I.~Frosio, and J.~Kautz, ``Loss functions for image restoration with neural networks,'' \emph{IEEE Transactions on computational imaging}, vol.~3, no.~1, pp. 47--57, 2016.

\bibitem{cho2021rethinking}
S.-J. Cho, S.-W. Ji, J.-P. Hong, S.-W. Jung, and S.-J. Ko, ``Rethinking coarse-to-fine approach in single image deblurring,'' in \emph{Proceedings of the IEEE/CVF international conference on computer vision}, 2021, pp. 4641--4650.

\bibitem{tu2022maxim}
Z.~Tu, H.~Talebi, H.~Zhang, F.~Yang, P.~Milanfar, A.~Bovik, and Y.~Li, ``Maxim: Multi-axis mlp for image processing,'' in \emph{Proceedings of the IEEE/CVF conference on computer vision and pattern recognition}, 2022, pp. 5769--5780.

\bibitem{islam2020fast}
M.~J. Islam, Y.~Xia, and J.~Sattar, ``Fast underwater image enhancement for improved visual perception,'' \emph{IEEE Robotics and Automation Letters}, vol.~5, no.~2, pp. 3227--3234, 2020.

\bibitem{loshchilov2016sgdr}
I.~Loshchilov and F.~Hutter, ``Sgdr: Stochastic gradient descent with warm restarts,'' \emph{arXiv preprint arXiv:1608.03983}, 2016.

\bibitem{liu2022rank}
J.~Liu, R.~W. Liu, J.~Sun, and T.~Zeng, ``Rank-one prior: Real-time scene recovery,'' \emph{IEEE Transactions on Pattern Analysis and Machine Intelligence}, 2022.

\bibitem{zhou2023underwater}
J.~Zhou, Q.~Liu, Q.~Jiang, W.~Ren, K.-M. Lam, and W.~Zhang, ``Underwater camera: Improving visual perception via adaptive dark pixel prior and color correction,'' \emph{International Journal of Computer Vision}, pp. 1--19, 2023.

\bibitem{zhang2023underwater}
W.~Zhang, L.~Zhou, P.~Zhuang, G.~Li, X.~Pan, W.~Zhao, and C.~Li, ``Underwater image enhancement via weighted wavelet visual perception fusion,'' \emph{IEEE Transactions on Circuits and Systems for Video Technology}, 2023.

\bibitem{li2022beyond}
K.~Li, L.~Wu, Q.~Qi, W.~Liu, X.~Gao, L.~Zhou, and D.~Song, ``Beyond single reference for training: underwater image enhancement via comparative learning,'' \emph{IEEE Transactions on Circuits and Systems for Video Technology}, 2022.

\bibitem{liu2022twin}
R.~Liu, Z.~Jiang, S.~Yang, and X.~Fan, ``Twin adversarial contrastive learning for underwater image enhancement and beyond,'' \emph{IEEE Transactions on Image Processing}, vol.~31, pp. 4922--4936, 2022.

\bibitem{ma2022wavelet}
Z.~Ma and C.~Oh, ``A wavelet-based dual-stream network for underwater image enhancement,'' in \emph{ICASSP}, 2022, pp. 2769--2773.

\bibitem{ren2022reinforced}
T.~Ren, H.~Xu, G.~Jiang, M.~Yu, X.~Zhang, B.~Wang, and T.~Luo, ``Reinforced swin-convs transformer for simultaneous underwater sensing scene image enhancement and super-resolution,'' \emph{IEEE Transactions on Geoscience and Remote Sensing}, vol.~60, pp. 1--16, 2022.

\bibitem{fu2022unsupervised}
Z.~Fu, H.~Lin, Y.~Yang, S.~Chai, L.~Sun, Y.~Huang, and X.~Ding, ``Unsupervised underwater image restoration: From a homology perspective,'' in \emph{AAAI}, vol.~36, 2022, pp. 643--651.

\bibitem{mu2023generalized}
P.~Mu, H.~Xu, Z.~Liu, Z.~Wang, S.~Chan, and C.~Bai, ``A generalized physical-knowledge-guided dynamic model for underwater image enhancement,'' in \emph{ACM Multimedia}, 2023, pp. 7111--7120.

\bibitem{huang2023contrastive}
S.~Huang, K.~Wang, H.~Liu, J.~Chen, and Y.~Li, ``Contrastive semi-supervised learning for underwater image restoration via reliable bank,'' in \emph{The IEEE/CVF Conference on Computer Vision and Pattern Recognition}, 2023, pp. 18\,145--18\,155.

\bibitem{wang2023domain}
Z.~Wang, L.~Shen, M.~Xu, M.~Yu, K.~Wang, and Y.~Lin, ``Domain adaptation for underwater image enhancement,'' \emph{IEEE Transactions on Image Processing}, vol.~32, pp. 1442--1457, 2023.

\end{thebibliography}

\end{document}